\documentclass[lettersize,journal]{IEEEtran}
\usepackage{amsmath,amsfonts}
\usepackage{algorithmic}
\usepackage{algorithm}
\usepackage{array}
\usepackage[caption=false,font=normalsize,labelfont=sf,textfont=sf]{subfig}
\usepackage{textcomp}
\usepackage{stfloats}
\usepackage{url}
\usepackage{verbatim}
\usepackage{graphicx}
\usepackage{cite}
\usepackage[numbers]{natbib}
\begin{document}

\title{Combining Microscopy Data and Metadata for Reconstruction of Cellular Traction Forces Using a Hybrid Vision Transformer–U-Net}


\author{Yunfei Huang\textsuperscript{1}, Elena Van der Vorst\textsuperscript{2,3}, Alexander Richard\textsuperscript{3}, Benedikt Sabass\textsuperscript{1,2,3*}
\thanks{\textbf{1}, Fakultät Physik, Technische Universität Dortmund, Dortmund, 44227, Germany. \textbf{2} Faculty of Physics and Center for NanoScience, Ludwig-Maximilians-Universit\"at M\"unchen, Munich, 80752, Germany. \textbf{3} Department of Veterinary Sciences, Ludwig-Maximilians-Universit\"at M\"unchen, Munich, 80752, Germany. }
\thanks{*benedikt.sabass@tu-dortmund.de}}

\markboth{Journal of \LaTeX\ Class Files,~Vol.~14, No.~8, August~2021}%
{Shell \MakeLowercase{\textit{et al.}}: A Sample Article Using IEEEtran.cls for IEEE Journals}


\maketitle

\begin{abstract}
Traction force microscopy (TFM) is a widely used technique for quantifying the forces that cells exert on their surrounding extracellular matrix. Although deep learning methods have recently been applied to TFM data analysis, several challenges remain—particularly achieving reliable inference across multiple spatial scales and integrating additional contextual information such as cell type to improve accuracy. In this study, we propose \textit{ViT+UNet}, a robust deep learning architecture that integrates a U-Net with a Vision Transformer. Our results demonstrate that this hybrid model outperforms both standalone U-Net and Vision Transformer architectures in predicting traction force fields. Furthermore, \textit{ViT+UNet} exhibits superior generalization across diverse spatial scales and varying noise levels, enabling its application to TFM datasets obtained from different experimental setups and imaging systems. By appropriately structuring the input data, our approach also allows the inclusion of metadata, in our case cell-type information, to enhance prediction specificity and accuracy.
\end{abstract}


\section{Introduction}
\IEEEPARstart{C}{ells} generate mechanical forces on their surrounding extracellular matrix (ECM) and neighboring cells. These cellular traction forces regulate fundamental biological processes such as migration~\cite{wolf2013physical,bohringer2024dynamic,hockenberry2025measurement}, wound healing~\cite{brugues2014forces,ajeti2019wound}, and immune responses~\cite{vorselen2020microparticle,reis2025migrating,basu2016cytotoxic}. Dysregulation of these forces has been implicated in various diseases ranging from cancer~\cite{koch20123d,aung20143d,barbazan2023cancer,du2025intercellular} to liver fibrosis~\cite{caliari2016stiffening,fan2024matrix}. Accurate quantification of these mechanical interactions is therefore essential for understanding cellular behavior under both physiological and pathological conditions.

\begin{figure*}[htp]
    \centering
    \includegraphics[width=\textwidth]{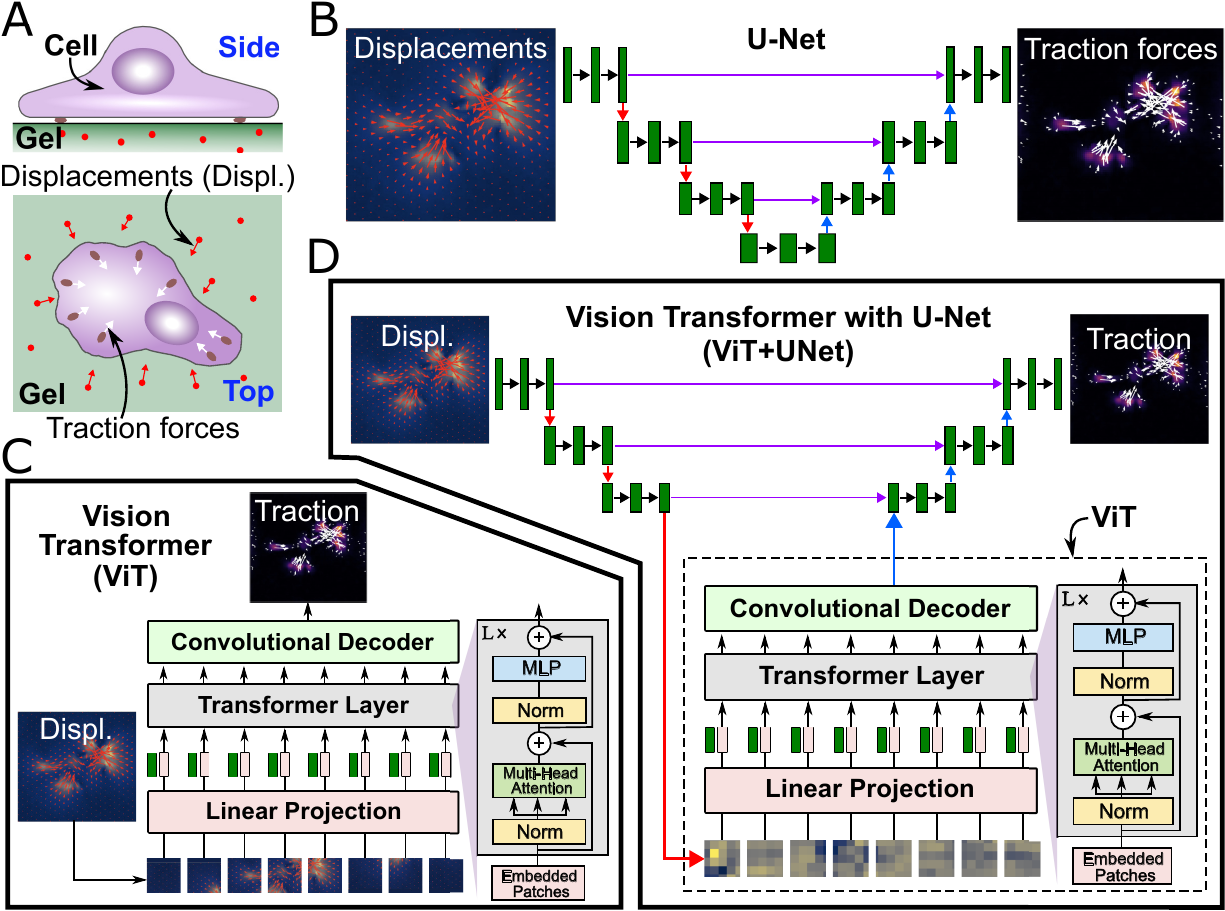}
   \caption{\textbf{Schematic diagram of traction force microscopy and the architectures of deep learning models used for traction force analysis.} 
(A)~Experimental setup: a single cell adheres to the surface of an elastic gel, and the traction force field is calculated from the measured displacement field. 
(B)~Architecture of U-Net, consisting of downsampling, a middle block, and upsampling stages. The black arrows indicate 2D convolutions; red arrows represent downsampling through 2D convolutions; blue arrows denote 2D transposed convolutions used for upsampling; and purple arrows indicate skip connections (copy operations). The input and output of the model correspond to the displacement field and traction force field, respectively. 
(C)~Architecture of Vision Transformer (ViT), which includes image pre-processing, a transformer encoder, and a convolutional decoder. Note that the transformer encoder contains $L$ layers. 
(D)~Architecture of the hybrid deep learning model \textit{ViT+UNet}, which combines a Vision Transformer with a U-Net framework. In this design, the middle block of U-Net is replaced by the Vision Transformer module.}
    \label{fig:architecture}
\end{figure*}

Traction force microscopy (TFM) is an effective and widely used technique for inferring cellular traction forces $\mathbf{f}$ on a substrate from measured displacement fields $\mathbf{u}$ based on the known mechanical properties of the substrate, as illustrated in Fig.~(\ref{fig:architecture}-A). The displacement field is typically obtained by tracking fluorescent marker beads embedded within the substrate using particle image velocimetry (PIV). The mechanical relationship between force and displacement can be numerically approximated using methods such as the boundary element method (BEM)~\cite{sabass2008high,huang2019traction}, finite element method (FEM)~\cite{yang2006determining,zielinski2013finite}, or Fourier-transform traction cytometry (FTTC)~\cite{sabass2008high,huang2020bayesian}. Determining unknown traction forces from measured displacements constitutes an inverse problem that is inherently ill-posed—small errors in measurement can lead to large instabilities in the reconstructed force fields. Classical approaches rely on regularization techniques such as $L_2$ regularization~\cite{schwarz2002calculation,sabass2008high}, elastic-net regularization~\cite{huang2019traction}, or non-parametric Bayesian regularization to stabilize solutions~\cite{huang2019traction,huang2020bayesian}. Despite these advances, traditional physics-based methods remain computationally expensive, limited to single deterministic predictions, and sensitive to experimental noise.

Recently, machine learning (ML) models have emerged as powerful tools for solving this inverse problem by learning direct mappings between displacement data and traction forces rather than relying solely on physical modeling with regularization techniques. ML-based approaches enable faster inference once trained while maintaining high predictive performance even under complex experimental conditions. Most existing deep-learning frameworks focus on two-dimensional (2D) traction or stress fields derived from displacement data~\cite{wang2021traction,kratz2023enhancing,li2024machine}. Furthermore, a trained 2D machine-learning model can infer both displacement and traction force fields directly from sequential cell images~\cite{tao2024inferring}. In another approach, a deep-learning model was first trained to learn surface wrinkles and subsequently used to train another model that predicts the traction force field from images or surface wrinkle patterns~\cite{li2022wrinkle}. All these trained models are based on measured displacement information as input to infer the traction force field. Because of the robust capability of neural networks to extract nonlinear patterns and hidden features that extend beyond direct or analytically tractable physical relationships, advanced architectures have also been proposed to directly infer traction forces from cellular protein images without explicit displacement information~\cite{schmitt2024machine}. Furthermore, three-dimensional deep-learning–based frameworks have been developed to capture out-of-plane forces and complex 3D cell–matrix interactions~\cite{duan2022deep,barrasa2025data}, enabling data-driven TFM analysis in three-dimensional collagen hydrogels. Although existing machine-learning models efficiently predict traction forces with high robustness across different datasets, they still face challenges arising from untested techniques, evolving algorithms, and newly developed datasets.

U-Net~\cite{ronneberger2015u} and Vision Transformers (ViT)~\cite{vit} are two prominent neural network architectures widely applied in image classification~\cite{cao2020improved,yang2023novel,borhani2022deep,bhojanapalli2021understanding} and segmentation tasks~\cite{huang2020unet,weng2019unet,strudel2021segmenter,zhang2022segvit}. U-Net excels at identifying short-range dependencies associated with local spatial patterns, whereas ViT specializes in capturing long-range dependencies related to global contextual information. Consequently, several advanced hybrid architectures have been developed that combine both approaches—including Swin-UNet~\cite{cao2022swin}, TransUNet~\cite{chen2024transunet}, SUNet~\cite{fan2022sunet}, META-UNet~\cite{wu2023meta}, and After-UNet~\cite{yan2022after}—for various image analysis tasks; however, whether these hybrid networks improve traction force prediction remains unclear. Moreover, ViT provides a robust architecture for integrating both image and text metadata inputs across different tasks—for example, image generation~\cite{chen2024pixart} and image segmentation~\cite{luddecke2022image}. Nevertheless, the use of such multimodal (image–text) metadata paired inputs in TFM applications remains largely unexplored.

Generalization—the ability of a model to perform well on unseen data—is an essential metric for evaluating machine-learning performance. It ensures applicability beyond the training dataset and is crucial for real-world biological imaging applications, where variability arises due to differences in experimental setups or imaging scales. In deep-learning-based TFM studies, generalization has been extensively discussed and investigated across different cell types and biochemical perturbations~\cite{schmitt2024machine}; however, achieving consistent performance across diverse spatial scales and varying noise levels remains challenging.


In this work, we introduce a hybrid neural network architecture (ViT+UNet) for traction force microscopy that combines the classical U-Net for capturing short-range dependencies with a robust Vision Transformer (ViT) for modeling long-range contextual relationships. We compare the performance of ViT+UNet with that of the individual U-Net and ViT models, as well as with Bayesian Fourier-transform traction cytometry (BFTTC)~\cite{huang2020bayesian}. Furthermore, we investigate the generalization capabilities of these models across diverse spatial scales and evaluate their predictive robustness under varying noise levels. In addition, we develop extended versions of ViT+UNet and ViT capable of inferring traction forces using metadata by incorporating both displacement fields and cell-type information.

\section{Materials and methods}
\subsection{U-Net}
The standard U-Net architecture is one of the most widely known convolutional neural networks and was originally developed for biomedical image segmentation~\cite{ronneberger2015u}. Recently, enhanced versions of U-Net have been proposed that integrate wide residual blocks with attention mechanisms throughout the network—across the downsampling, middle, and upsampling stages—collectively referred to as modern U-Net~\cite{huang2025geometric}. In this work, we present a modified version of the modern U-Net in which attention mechanisms are removed while residual blocks are retained within the architecture, as illustrated in Fig.~(\ref{fig:architecture}-B).

\textbf{Downsampling.} The input features processed by the U-Net are represented by the displacement field $\mathbf{u}\in \mathbb{R}^{2\times N \times N}$, where the first dimension ($2$) corresponds to the two components of the displacement field, $u_x$ and $u_y$. The parameter $N$ denotes the image size of the displacement field, which is set to $104$ in this work. The downsampling process consists of three stages, each comprising a {ResidualBlock} (indicated by black arrows) and a {Downsample} operation (indicated by red arrows). The {ResidualBlock} is implemented using Wide Residual Blocks~\cite{zagoruyko2016wide}. In each block, we add the input to two convolutional layers with GELU activation and group normalization. The output of the {ResidualBlock}, $y$, is defined as  
\begin{equation}
y = x + 2\times \text{Conv}\big(\text{GELU}(x)\big),
\label{eq:ResidualBlock}
\end{equation}
where $x$ denotes the input. The GELU activation function is defined as $\text{GELU}(x) = x\,\Phi(x)$, with $\Phi(x)$ representing the cumulative distribution function of a standard normal distribution. The \textit{Downsample} operation uses a convolutional layer to reduce the spatial dimensions of the feature map by a factor of two.

\textbf{Middle block.} The middle block at the lowest resolution level of the U-Net consists of a single {ResidualBlock}.

\textbf{Upsampling.} The upsampling process also consists of three stages, each containing a ResidualBlock and an upsampling operation (indicated by blue arrows). The ResidualBlock is defined exactly as in Eq.~(\ref{eq:ResidualBlock}). A  ConvTranspose2D layer is used for upsampling after each {ResidualBlock}. Skip connections (indicated by purple arrows) are established between the corresponding downsampling and upsampling layers to preserve spatial information. At the final upsampling stage, the traction force field $\mathbf{f}\in \mathbb{R}^{2\times N\times N}$ is produced as the network output.

\subsection{Vision Transformer}

\textbf{Image Pre-Processing.} The purpose of image pre-processing is to convert the input images into a sequence of patches. The displacement field for the Vision Transformer is represented as $\mathbf{u}\in \mathbb{R}^{2\times N\times N}$, as shown in Fig.~(\ref{fig:architecture}-C). The image pre-processing includes the following steps:
\begin{itemize}
    \item \textbf{Image Splitting:} The ViT divides the input into $K$ non-overlapping patches of size $2\times P\times P$, such that $K = N^2/P^2$.
    \item \textbf{Flattening:} Each patch is flattened into a vector of shape $2P^2$.
    \item \textbf{Linear Embedding:} Each flattened patch is projected into a $D$-dimensional embedding space through a learnable linear transformation. This operation allows the model to represent each patch as a high-dimensional feature vector that can be analyzed relative to other patches.
    \item \textbf{Positional Encoding:} To preserve spatial information, learnable positional encodings are added to each patch embedding. These encodings provide information about the relative positions of patches within the displacement field, enabling the model to capture both local structures and long-range spatial relationships.
\end{itemize}

\textbf{Transformer.} The sequence of patch embeddings is represented as a matrix $z\in \mathbb{R}^{K\times D}$, where each row corresponds to one image patch. This representation is then passed through a transformer encoder consisting of $L$ layers. Each layer includes two main components: a multi-head self-attention (MSA) block and a feed-forward network (MLP). An overview of this structure is illustrated in Fig.~(\ref{fig:architecture}-C). The MSA mechanism forms the core component of the transformer encoder; it enables each patch to integrate information from all others, thereby identifying regions most relevant to its representation. Analogous to long-range dependencies in language models, localized traction may induce stress propagation across the field, rendering distant regions informative.
\begin{itemize} 
\item \textbf{Multi-head Self-attention (MSA):} Each patch embedding $z_i \in \mathbb{R}^D$ is linearly projected into query, key, and value vectors:
\begin{equation}
\overrightarrow{q_i} = W_Qz_i, \quad \overrightarrow{k_i} = W_Kz_i, \quad  \overrightarrow{v_i} = W_Vz_i,
\end{equation}
where $W_Q, W_K, W_V \in \mathbb{R}^{D\times d_k}$ are learnable projection matrices. The attention weight between embeddings $q_i$ and $k_j$  is computed using scaled dot-product attention: 
\begin{equation}
\text{weight}_{i,j}=\text{softmax}\Bigg(\frac{\overrightarrow{q_i}\cdot\overrightarrow{k_j}}{\sqrt{d_k}}\Bigg). 
\end{equation}
These weights determine the contribution of each patch $k_j$  to the representation of patch $q_i$. The output representation for each patch is obtained as a weighted sum over the value vectors:
\begin{equation}
y_i=\sum_j(\text{weight}_{i,j}\times \overrightarrow{v_i}),
\end{equation}
yielding contextualized embeddings that integrate information from both local and distant regions.  The MSA mechanism employs  $h$ parallel attention heads with independent projections. Their outputs are concatenated and linearly mapped back to the original embedding dimension, integrating multiple contextual perspectives into a unified representation. The output $y$ is connected from input $x$ by using  residual connection
\begin{equation}
y =x + \text{MSA}(\text{Dropout}(x)),
\end{equation}

\item  \textbf{Multi-Layer Perceptron (MLP):} Following the MSA module, each transformer layer includes an MLP module that processes each patch embedding independently. Formally, the MLP consists of two linear layers separated by a non-linear activation function (GELU):
\begin{equation}
\text{MLP}(x) = 2\times \text{Dropout}\big(\text{GELU}(W_1 \cdot \text{LayerNorm}(x) + b_1)\big),
\label{eq:mlp}
\end{equation}
where $W_1$ and $b_1$ are learnable weight and bias parameters. The GELU introduces non-linearity into the MLP, allowing the model to capture complex interactions between features within each patch. The output $y$ from the MLP is connected to its input through a residual connection:
\begin{equation}
y = x + \text{MLP}(x),
\end{equation}
\end{itemize}

\begin{figure*}[htp]
\centering
\includegraphics[width=1\textwidth]{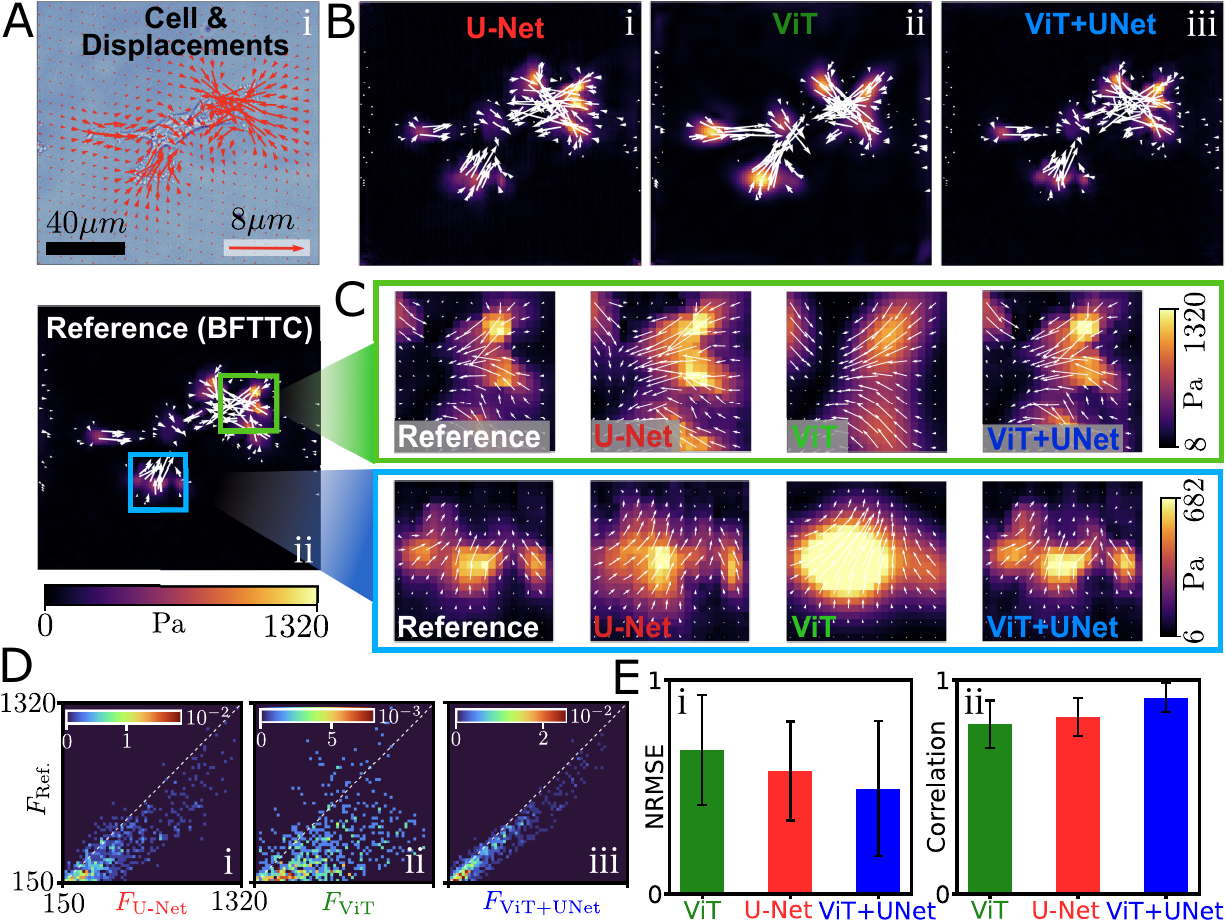}
\caption{\textbf{The hybrid \textit{ViT+UNet} architecture consistently outperforms both standalone U-Net and Vision Transformer models in predicting traction force fields.} 
(A)~Representative cell image and its corresponding displacement field used as input (i). The reference traction force field, calculated using Bayesian Fourier-transform traction cytometry (BFTTC), is shown in (ii). Note that the vector field is plotted only for regions where the traction magnitude exceeds $50\,\mathrm{Pa}$, skipping every 15 indices for clarity. 
(B, i–iii)~Comparison of predicted traction force fields obtained from U-Net, Vision Transformer (ViT), and \textit{ViT+UNet} models. 
(C)~Zoomed-in views corresponding to the colored boxes in panel (A-ii). 
(D, i–iii)~Joint probability distributions of force magnitudes for the predictions and references, $p(F_\text{Ref}, F_\text{NN})$, obtained from U-Net, ViT, and \textit{ViT+UNet} models. Forces larger than $150\,\mathrm{Pa}$ are considered; better predictions with higher probability density values lie along the diagonal dotted line. 
(E)~Statistical comparison of normalized root mean square error (NRMSE; i) and correlation coefficients (ii) across 34 test cases. These two metrics are defined in Eqs.~(\ref{eq:NRMSE}) and (\ref{eq:correlation}). Lower NRMSE values indicate more accurate predictions, while correlation coefficients closer to 1 reflect stronger agreement with the reference data. Error bars represent standard deviations.}
\label{fig:vit_unet}
\end{figure*}

\textbf{Convolutional Decoder.} The output from the contextualized patch representations $y \in \mathbb{R}^{K\times D}$ produced by the transformer is passed to a convolutional decoder to reconstruct the spatial traction force field $\mathbf{f}\in \mathbb{R}^{2\times N\times N}$. We follow Ref.~\cite{zheng2021rethinking} for the design of the convolutional decoder following the transformer. The output matrix is first reshaped to ${D\times (N/P \times N/P)}$. Subsequently, three convolutional neural network layers are employed for upsampling, each using GELU activation functions, to produce an output of size $2\times N\times N$, corresponding to the reconstructed traction force field.

\subsection{Hybrid Model: ViT+UNet}
In this work, we propose a new ViT+UNet architecture that combines the U-Net with a Vision Transformer (ViT), as illustrated in Fig.~(\ref{fig:architecture}-D). This approach is similar to TransUNet, which was originally developed for medical image segmentation~\cite{chen2024transunet}. In our design, the ViT component captures long-range dependencies and global contextual information, while the U-Net effectively models short-range dependencies and local spatial patterns. Specifically, the middle block at the lowest resolution level of the U-Net is replaced by a standard Vision Transformer. In this implementation, the input feature map passed from the downsampling stage of the U-Net to the ViT has dimensions $512\times13\times13$. These internal representation features are directly processed by the standard ViT, where they are divided into patches of size $512\times1\times1$. At the end of the ViT module, the output from its convolutional decoder has dimensions $512\times13\times13$, which is subsequently fed into the upsampling stage of the U-Net.

\begin{table*}[h!]
\centering
\renewcommand{\arraystretch}{1.3} 
\begin{tabular}{cccc}
\noalign{\hrule height 0.29mm} 
Model & Parameters of network & Initial training rate & Training time per epoch   \\ 
\hline
U-Net   &26536178  & 0.0002  &  $7.056\,\mathrm{s}$ \\ 
ViT     &24490148 & 0.0002   & $7.485\,\mathrm{s}$ \\
ViT+UNet &24357546  & 0.0002 &  $9.679\,\mathrm{s}$ \\
\noalign{\hrule height 0.29mm} 
\end{tabular}
\caption{Overview of the architecture and training parameters for the U-Net, Vision Transformer (ViT), and ViT+UNet.}
\label{table1}
\end{table*}

\subsection{Training}
\label{sect:training}

\textbf{Dataset.} We used 225 individual real-cell cases as training data. The input training dataset consists of measured displacement fields with dimensions $225\times2\times104\times104$. The corresponding traction forces were computed from these displacement fields using Bayesian Fourier-transform traction cytometry (BFTTC)~\cite{huang2020bayesian}, based on the Poisson ratio and Young’s modulus of the substrate. Consequently, the training dataset comprises paired displacement and traction force fields with dimensions $225\times2\times104\times104$. For model evaluation, 34 real-cell cases were employed for testing and 61 for validation, resulting in datasets of sizes $34\times2\times104\times104$ and $61\times2\times104\times104$, respectively.

\textbf{Loss Function.} A supervised learning approach is employed to train the models. Accordingly, we use the mean squared error (MSE) loss function, defined as
\begin{equation}
\mathcal{L} = \frac{1}{2N^2}\sum_{i=1}^{2N^2}(f_i - \hat{f}_i)^2,
\label{eq:loss}
\end{equation}
where $\hat{f}_i$ denotes the predicted traction force from the neural network model and $f_i$ represents the target (observed) force. Both the predicted and target traction forces are represented as $2N^2\times1$ vectors reshaped from $\mathbf{\hat{f}}$ and $\mathbf{f}$, with $N=104$.

We implemented three models—U-Net, Vision Transformer (ViT), and ViT+UNet—using PyTorch and NumPy. The details of the model sizes are summarized in Table~(\ref{table1}). The initial learning rate was set to $0.0002$ for all models, and the \texttt{StepLR} function was used to decay the learning rate of each parameter group by a factor of $\gamma = 0.9$ every 40 epochs. Early stopping was applied when the validation loss did not decrease for 10 consecutive epochs, and the network weights corresponding to the best validation loss were retained throughout training. Furthermore, we utilized the Distributed Data Parallel (DDP) strategy for efficient multi-GPU training on two NVIDIA A100-PCIE-40GB GPUs. The average training time per epoch for U-Net, ViT, and ViT+UNet was $7.056\,\mathrm{s}$, $7.485\,\mathrm{s}$, and $9.679\,\mathrm{s}$, respectively.

\subsection{Evaluation Metrics}
To evaluate the quality of traction force predictions from the deep learning models, we introduce three evaluation metrics that compare the predicted traction forces $\mathbf{\hat{f}} =(\hat{f}_x, \hat{f}_y)$ with their corresponding reference (BFTTC) values $\mathbf{f}=(f_x, f_y)$  from the test data. The magnitudes of the resultant forces for the predictions and references are represented as  $N^2\times 1$ vectors and denoted respectively as $F_\text{NN}$ and $F_\text{Ref.}$. Each component of these vectors is defined as $f_i^{\text{NN}} =\sqrt{(\hat{f}_x)_i^2+(\hat{f}_y)_i^2} $ and $ f_i^{\text{Ref.}} =\sqrt{(f_x)_i^2+(f_y)_i^2} $.


\begin{itemize}
    \item \textbf{Normalized Root Mean Square Error (NRMSE).} We quantify the relative difference between the predicted traction forces and the reference traction forces as follows:
    \begin{align}
        \text{NRMSE} &= \frac{\left\lVert F_\text{NN} - F_\text{Ref.}\right\rVert_2}{\left\lVert F_\text{Ref.}\right\rVert_2} \label{eq:NRMSE}. 
     \end{align} 
     Note that the range of \text{NRMSE} extends from zero to infinity, where lower values indicate better predictions, with the best results corresponding to \text{NRMSE} values close to zero.
    
    \item \textbf{Pearson’s Correlation Coefficient.} We employ the Pearson correlation coefficient, $\rho(\mathbf{\hat{f}}, \mathbf{f})$, for each component of the predicted and reference traction forces to quantify their linear relationship.
    \begin{align}
    \rho(\mathbf{\hat{f}}, \mathbf{f}) = \frac{\sum^{2N^2}_{i=1}(f_i-\Bar{f})(\hat{f}_i-\Bar{\hat{f}})}{\sqrt{\sum^{2N^2}_{i=1}(f_i-\Bar{f})^2} \sqrt{\sum^{2N^2}_{i=1}(\hat{f}_i-\Bar{\hat{f}})^2}}, \label{eq:correlation}
    \end{align}
     where $\Bar{\hat{f}}$ and $\Bar{f}$ denote the mean traction values of each component of the predicted and reference traction forces, respectively. The range of $\rho(\mathbf{\hat{f}}, \mathbf{f})$ is from –1 to 1, with values close to 1 indicating better predictions.
    
    \item \textbf{Joint Probability Distributions of Force Magnitudes.} We compute the conditional distributions of the magnitudes of the predicted and reference traction forces, $p(F_\text{Ref}, F_\text{NN})$, using normalized 2D histogram binning over the same range of resultant force magnitudes. A similar metric based on conditional distributions between predicted and reference traction forces in TFM has been reported in Ref.~\cite{schmitt2024machine}. The most accurate predictions occur when $p(F_\text{Ref}, F_\text{NN})$ is concentrated along the diagonal line, indicating that the predicted and reference traction forces exhibit similar magnitudes across corresponding ranges.
\end{itemize}

\section{Results}
\subsection{The hybrid ViT+UNet architecture consistently outperforms both standalone U-Net and Vision Transformer models in predicting traction force fields}  

To qualitatively assess the performance of the proposed ViT+UNet model on real experimental data, we compared it with standalone U-Net~\cite{huang2025geometric} and Vision Transformer (ViT)~\cite{vit} architectures of comparable network size (see Table~\ref{table1}). All models were trained using the same dataset and training parameters, as described in Section~\ref{sect:training}. Fig.~(\ref{fig:vit_unet}-A) shows a representative cell image together with its corresponding displacement vector field used for inference (i) and the reference traction force field obtained using Bayesian Fourier-transform traction cytometry (BFTTC), illustrating both magnitude and vector directions (ii), taken from one of the 34 test cases.

\begin{figure*}[htp]
\centering
\includegraphics[width=1\textwidth]{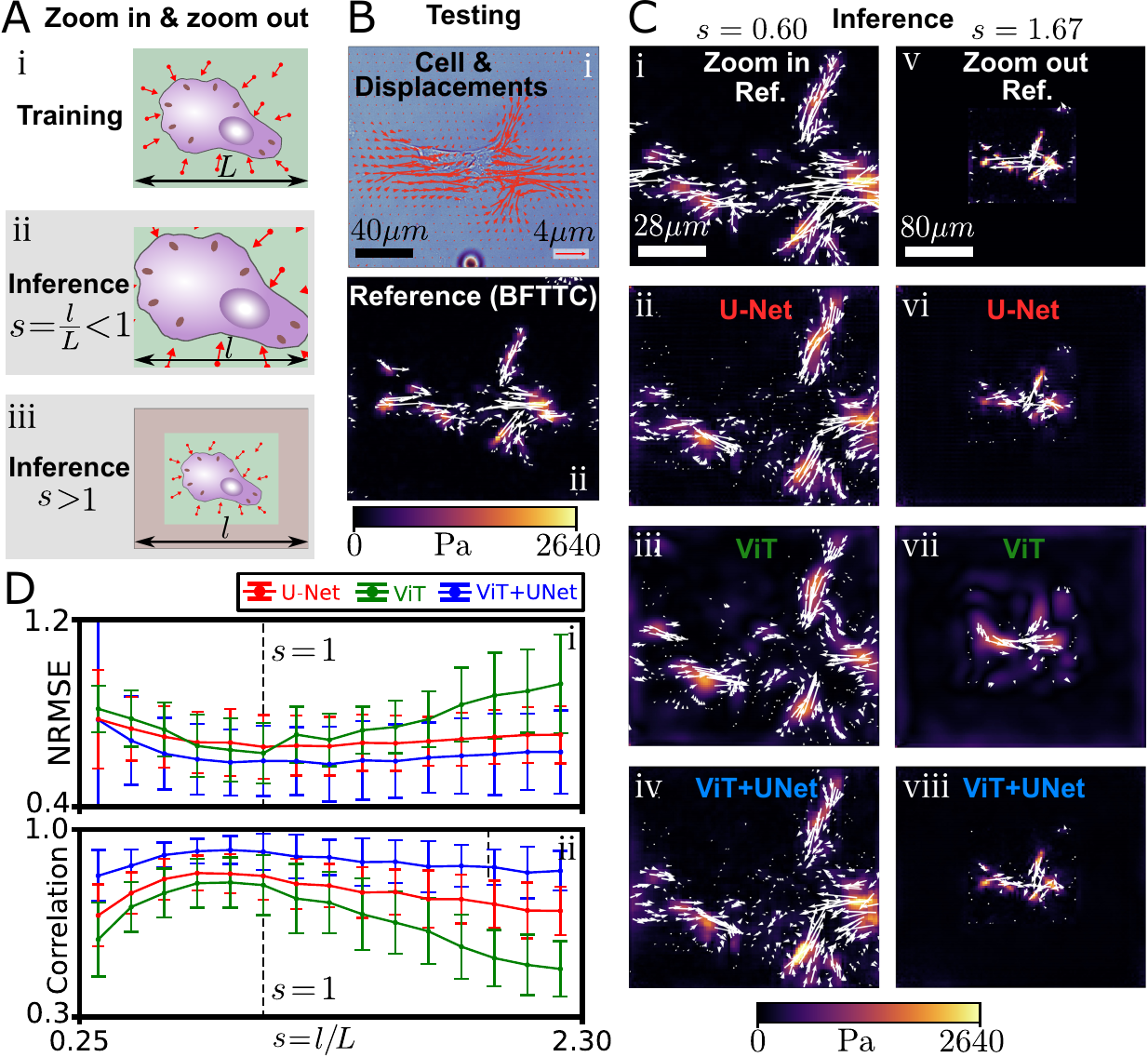}
\caption{\textbf{The \textit{ViT+UNet} architecture exhibits superior generalization across diverse spatial scales.} 
(A)~Schematic diagram illustrating the inference procedure for different spatial scales. The scale ratio $s$ is defined as $s = l / L$, where $l$ is the scaled length used during inference and $L$ is the original length used during training, as shown in (i). For inference cases with $s < 1$ and $s > 1$, these are referred to as zoom-in and zoom-out scenarios, respectively, illustrated in (ii) and (iii). 
(B)~Example of an original cell image with its corresponding displacement field (i) and traction force field obtained using Bayesian Fourier-transform traction cytometry (BFTTC) in (ii). 
(C)~Generalization inference results for the zoom-in ($s = 0.6$; ii–iv) and zoom-out ($s = 1.67$; vi–viii) scenarios obtained from U-Net, ViT, and \textit{ViT+UNet} models compared with the reference data shown in panels (i, v). 
(D)~Statistical summary showing normalized root mean square error (NRMSE; i) and correlation coefficients (ii) for \textit{ViT+UNet}, U-Net, and ViT across 34 test cases over scale ratios ranging from $s = 0.25$ to $2.3$. Error bars represent standard deviations.}
\label{fig:spatial_scale}
\end{figure*}

For the given displacement field shown in Fig.~(\ref{fig:vit_unet}-A, i), the traction forces were predicted using the U-Net, ViT, and ViT+UNet models, as illustrated in Fig.~(\ref{fig:vit_unet}-B, i–iii). We observe that the inference pattern produced by ViT+UNet closely matches the reference shown in Fig.~(\ref{fig:vit_unet}-A, ii) in terms of spatial distribution, shape, magnitude, and direction of force when compared with the other two models (U-Net and ViT; see panels i and ii). However, the prediction obtained from U-Net exhibits a similar overall shape but tends to overestimate both the spatial domain and magnitude relative to the reference traction force. Moreover, the directions of forces predicted by U-Net deviate from those of the reference in several local regions. The results obtained from ViT capture only large-scale features while failing to reproduce fine local details accurately compared with the reference; furthermore, its predicted magnitudes, spatial domains, shapes, and force directions are inferior to those produced by U-Net and ViT+UNet. These differences can be clearly observed in the zoomed-in views shown in Fig.~(\ref{fig:vit_unet}-C), which correspond to the two colored green and blue boxes indicated in Fig.~(\ref{fig:vit_unet}-A, ii).

To evaluate the accuracy of these models, we first quantitatively compared the joint probability distributions of force magnitudes between the predicted and reference traction forces, $p(F_\text{Ref}, F_\text{NN})$, obtained from U-Net, ViT, and ViT+UNet, as illustrated in Fig.~(\ref{fig:vit_unet}-D, i–iii). These distributions were plotted using normalized 2D histogram binning for resultant force magnitudes ranging from $150\,\mathrm{Pa}$ to $1320\,\mathrm{Pa}$ in the case shown in Fig.~(\ref{fig:vit_unet}-B), since our analysis primarily focuses on high-force regions. We found that the distribution of ViT, $p(F_\text{Ref}, F_\text{ViT})$, is most widely spread on both sides of the diagonal dotted line in Fig.~(\ref{fig:vit_unet}-D, ii). The distribution obtained from U-Net, $p(F_\text{Ref}, F_\text{U-Net})$, is narrower than that of ViT, as illustrated in Fig.~(\ref{fig:vit_unet}-D, i). In contrast, the distribution produced by ViT+UNet is the narrowest and lies closest to the diagonal dotted line in Fig.~(\ref{fig:vit_unet}-D, iii), indicating that the prediction distribution of ViT+UNet aligns most closely with the reference compared with the other two models.

Secondly, we evaluated two statistical metrics—the normalized root mean square error (NRMSE) and the correlation coefficient $\rho(F_\text{NN}, F_\text{Ref})$—computed over all 34 test cases, as illustrated in Fig.~(\ref{fig:vit_unet}-E, i–ii). The ViT+UNet model achieved the smallest NRMSE value ($0.493\pm0.314$), compared with U-Net ($0.574\pm0.231$) and ViT ($0.672\pm0.256$). Moreover, ViT+UNet yielded the highest correlation with the reference data ($0.917\pm0.068$), whereas U-Net and ViT achieved correlations of $0.827\pm0.088$ and $0.793\pm0.112$, respectively.

\subsection{The ViT+UNet Architecture Exhibits Superior Generalization Across Diverse Spatial Scales}
\label{section:vit+unet}

Recently, multiscale light-sheet imaging has been widely used to study the dynamics of tissue regeneration and development~\cite{de2022multiscale,yamauchi2022protocol}. An important feature of this technique is its ability to automatically acquire images across different spatial scales. In the context of TFM, these multiscale images can be utilized to calculate displacement fields at different spatial resolutions using particle image velocimetry (PIV). This raises two key questions concerning our deep-learning models for TFM:  
(1) Can trained neural network models for TFM automatically generalize to predict traction forces across varying spatial scales?  
and (2) Which architecture performs best in achieving this challenging task?

To address these two questions, we designed a series of tests to evaluate model generalization across diverse spatial scales. All deep-learning models were trained on paired input (displacement) and output (traction force) fields with dimensions $104\times104$, corresponding to a rectangular region of length $L$, as illustrated in Fig.~(\ref{fig:spatial_scale}-A, i). To test generalization across different spatial scales, we first reduced the length of the displacement (input) region from $L$ to $l<L$ and introduced a scale ratio defined as $s = l/L$. This process is referred to as ``zoom-in'' when $s < 1$, as illustrated in Fig.~(\ref{fig:spatial_scale}-A, ii). Secondly, we increased the length of the displacement (input) region from $L$ to $l > L$; this process is referred to as ``zoom-out'' when $s > 1$, as shown in Fig.~(\ref{fig:spatial_scale}-A, iii). For the zoom-in case, we cropped the displacement images and then interpolated them to a size of $104\times104$. For the zoom-out case, we applied zero padding and similarly interpolated them to a size of $104\times104$. It should be noted that these are generalization inference tests; therefore, the models used are identical to those described in the previous section, and only the displacement inputs are modified by spatial scaling during inference.

Figure~(\ref{fig:spatial_scale}-B) shows one of the test cell images and its corresponding displacement field in panel (i), while the traction force field calculated using BFTTC is shown in panel (ii). The size of this test case is referred to as the original size of the displacement and force fields during training. The displacement fields were tracked from images with a resolution of $1182\,\text{pix} \times 1042\,\text{pix}$, where each pixel corresponds to $0.161\,\mu\mathrm{m}/\text{pix}$. Thus, the original side length $L$ equals $190.3\,\mu\mathrm{m}$. First, we reduced $L$ to a spatial scale of $l = 113.45\,\mu\mathrm{m}$, corresponding to a scale ratio of $s = 0.6$, and interpolated it to a size of $104\times104$. The inference results obtained from all models for this ``zoom-in'' case are shown in Fig.~(\ref{fig:spatial_scale}-C, ii–iv). Next, we increased the spatial scale to $l = 318.39\,\mu\mathrm{m}$, corresponding to a scale ratio of $s = 1.67$, and similarly interpolated it to a size of $104\times104$; the results obtained from all models for this ``zoom-out'' case are presented in Fig.~(\ref{fig:spatial_scale}-C, vi–viii).

For the ``zoom-in'' case, when compared with the reference traction field in Fig.~\ref{fig:spatial_scale}(C, i), we observe that predictions from ViT+UNet and U-Net exhibit similar patterns in shape and force direction relative to the reference, as illustrated in Fig.~\ref{fig:spatial_scale}(C, ii) and (iv), respectively. However, the force magnitudes predicted by ViT+UNet are closer to those of the reference than those predicted by U-Net. In contrast, predictions from ViT (iii) are inferior to those from ViT+UNet and U-Net in terms of shape consistency, force magnitude accuracy, and directional alignment. 

For the ``zoom-out'' case, when compared with the reference shown in Fig.~\ref{fig:spatial_scale}(C, v), we can clearly observe that the pattern produced by ViT+UNet (viii) better matches the reference in terms of shape, force magnitude, and direction compared with U-Net (vi). In contrast, ViT (vii) performs noticeably worse than both models regarding shape fidelity, force magnitude accuracy, and directional coherence; additionally, its output exhibits a high level of background noise.

Finally, the plots of the statistical metrics—NRMSE and correlation coefficients—versus the scale ratio $s$ ranging from 0.25 to 2.3 across all 34 test cases are shown in Fig.~\ref{fig:spatial_scale}(D, i–ii), respectively. We observe that our ViT+UNet model (blue lines) outperforms both the individual U-Net and ViT architectures across all spatial scales, achieving the smallest NRMSE values and the highest correlation coefficients. The U-Net model (red line) yields smaller NRMSE values than the ViT model (green line), and its correlation coefficients are consistently higher than those of ViT. Furthermore, we find that the optimal prediction occurs at a scale ratio close to $s = 1$ for all models; beyond this point, prediction accuracy deteriorates as $s$ either increases or decreases.

\subsection{ViT+UNet demonstrates robust generalized prediction performance under different additional noise levels}

\begin{figure*}[htp]
    \centering
    \includegraphics[width=1.0\textwidth]{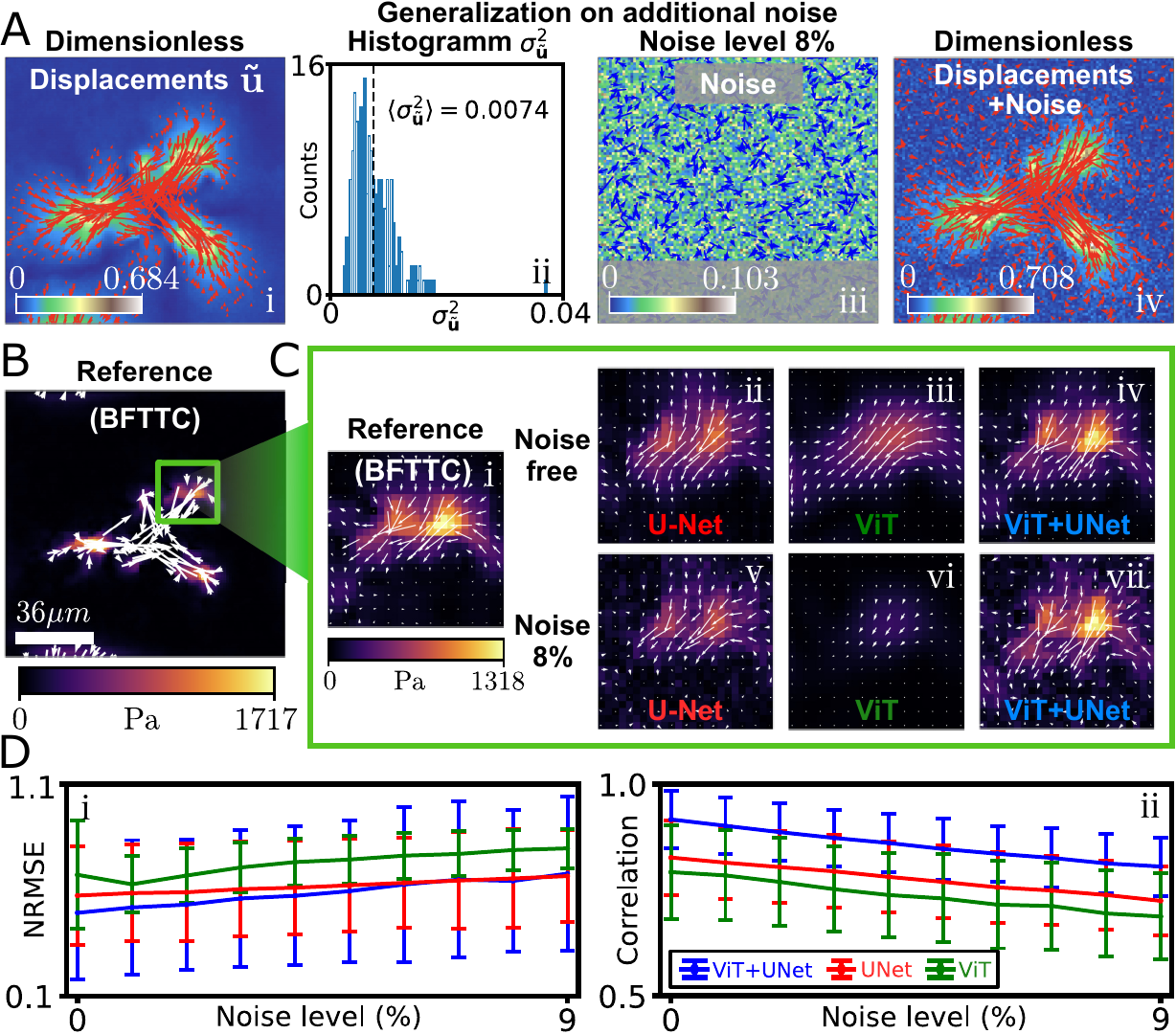}
    \caption{\textbf{The \textit{ViT+UNet} model generalizes well in predicting traction forces from displacement data under varying noise levels.}
(A-i)~Dimensionless displacement field showing both magnitude (color map) and vector directions. 
(A-ii)~Histogram of variance values and their average computed from dimensionless displacement fields in the training dataset. 
(A-iii)~Gaussian noise magnitude and corresponding vector field at an 8\% noise level. 
(A-iv)~Dimensionless displacement field after adding 8\% Gaussian noise, illustrating both magnitude (color map) and vector directions. 
(B)~Reference traction force field obtained using Bayesian Fourier-transform traction cytometry (BFTTC); colors indicate traction force magnitudes. 
(C)~Zoomed-in views of traction force fields under noise-free conditions [(ii–iv)] and at an 8\% noise level [(v–vii)] predicted by three models—U-Net, ViT, and \textit{ViT+UNet}—corresponding to the green box in panel~(B), compared with the reference region shown in (i). 
(D)~Plots of normalized root mean square error (NRMSE; i) and correlation coefficients (ii) for all three models at different noise levels ranging from 0 to 9\%. Error bars represent standard deviations.}
    \label{fig:noise}
\end{figure*}

To evaluate the generalization performance of our deep learning models in predicting traction forces under additional noise conditions, we introduced Gaussian noise into the input data during inference, while all deep learning models were trained on noise-free data. 
The Gaussian noise field has the same dimensions as the displacement field and consists of random numbers drawn from a normal distribution with mean~0 and variance~$\bar{\sigma}^2$. 
To define the noise level, we computed the average variance of the dimensionless displacement fields $\tilde{\mathbf{u}}$ in the training dataset, denoted by $\langle\sigma_{\tilde{\mathbf{u}}}^2\rangle$, where $\langle \cdot \rangle$ indicates averaging over all training samples. 
A representative dimensionless displacement field showing both magnitude (color map) and vector directions from the test dataset is presented in Fig.~\ref{fig:noise}(A-i). 
The histogram and average variance of the dimensionless displacement fields in the training data are shown in Fig.~\ref{fig:noise}(A-ii), where $\langle\sigma_{\tilde{\mathbf{u}}}^2\rangle = 0.0074$. 
An 8\% noise level was then defined according to Ref.~\cite{kratz2023enhancing} as
\begin{equation}
\bar{\sigma}^2 = 0.08\,\langle\sigma_{\tilde{\mathbf{u}}}^2\rangle,
\label{eq:noise_level}
\end{equation}
where $\bar{\sigma}^2$ denotes the variance of the added Gaussian noise. 
A Gaussian noise field at an 8\% level with the same size as the displacement field was generated and is shown in Fig.~\ref{fig:noise}(A-iii). The color map represents the magnitude of the noise, while vectors indicate its direction. The resulting noisy dimensionless displacement field after adding this 8\% Gaussian noise is shown in Fig.~\ref{fig:noise}(A-iv).

Figure~\ref{fig:noise}(B) shows the reference traction force field calculated using Bayesian Fourier-transform traction cytometry (BFTTC). 
Comparisons of predictions obtained from U-Net, ViT, and ViT+UNet—as well as reference data under both noise-free conditions and an 8\% noise level for the small green region indicated in Fig.~\ref{fig:noise}(B)—are presented in Fig.~\ref{fig:noise}(C). 
Under noise-free conditions, ViT+UNet provides excellent agreement with reference data in terms of traction magnitude, pattern shape, and force direction; U-Net performs second best; whereas ViT yields comparatively poor results among all three models. 
At an 8\% noise level, ViT+UNet maintains a prediction quality similar to its performance under noiseless conditions with respect to traction magnitude, pattern shape, and force direction [Fig.~\ref{fig:noise}(C-vii)]. For U-Net, predicted magnitudes become smaller than those obtained under noiseless conditions [Fig.~\ref{fig:noise}(C-v)], while ViT produces significantly lower magnitudes compared with its corresponding noiseless results [Fig.~\ref{fig:noise}(C-vi)]. Overall, ViT+UNet demonstrates superior generalization capability under noisy conditions while maintaining high-quality inference.

Next, we plotted statistical metrics—NRMSE and correlation coefficients—against different noise levels ranging from 0 to 9\,percent based on 34 test cases for all deep learning models (U-Net, ViT, and ViT+UNet), as shown in Fig.~\ref{fig:noise}(D-i, ii), respectively.
We observed that NRMSE increases for all models as the added noise level rises.
Furthermore, ViT+UNet consistently yields smaller NRMSE values than U-Net or ViT when subjected to less than six percent additional noise,
indicating superior generalization performance.
When exceeding six percent,
the NRMSE of ViT+UNet becomes slightly larger than that of U‑Net but remains smaller than that of ViT.
For correlation coefficients,
values decrease across all models with increasing amounts of added input perturbation.
ViT+UNet maintains a high correlation between its predictions and reference data across every tested condition—the blue line shown in Fig.~(\ref{fig:noise})(D‑ii)—followed by U‑Net,
while ViT exhibits consistently weaker correlations throughout.

\begin{figure*}[htp]
    \centering
    \includegraphics[width=1.0\textwidth]{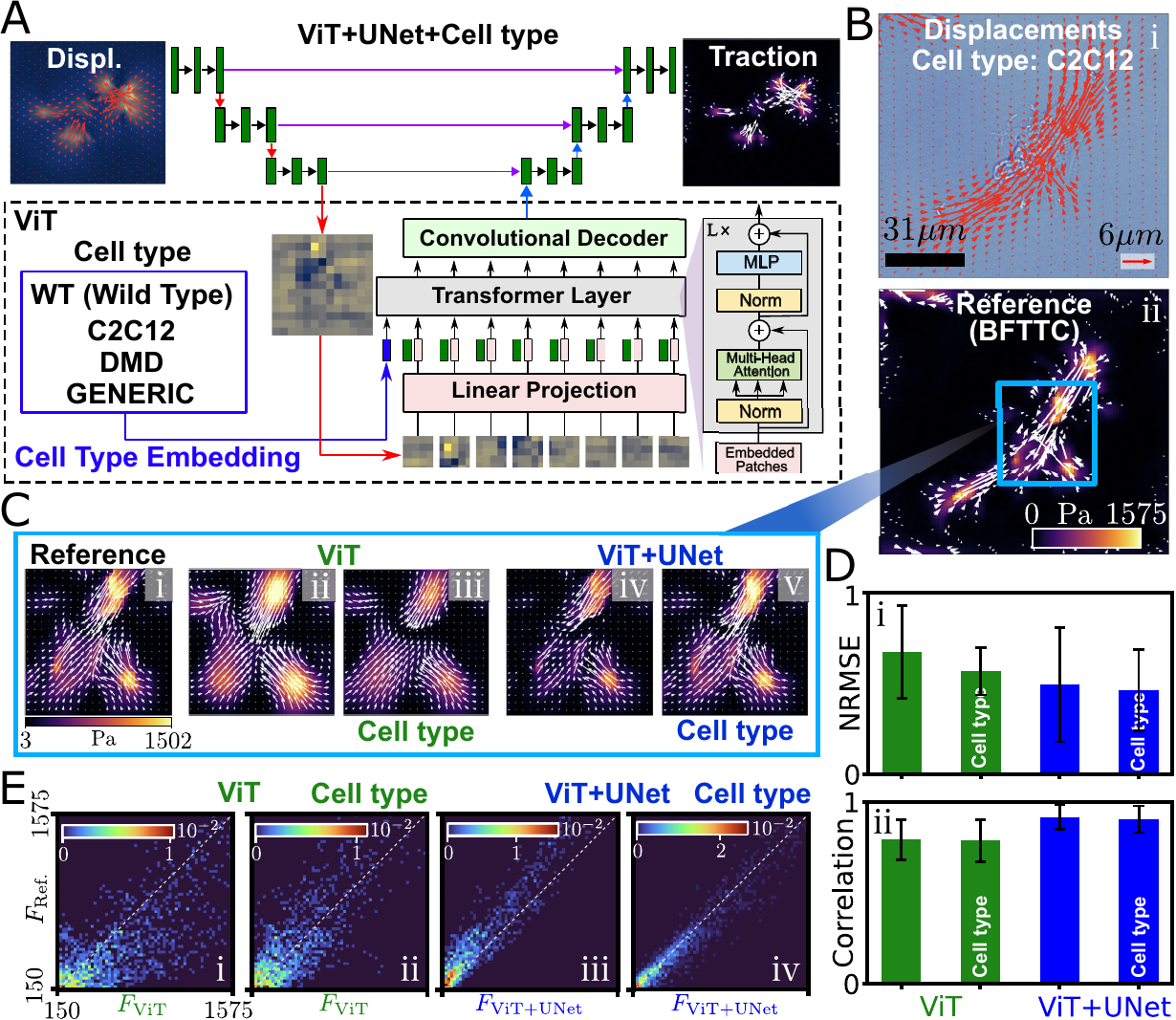}
\caption{\textbf{Incorporating cell-type information improves prediction specificity and accuracy.} 
(A)~Architecture of the hybrid deep learning model \textit{ViT+UNet}, which can also embed cell-type text information as an additional input. 
(B)~Representative cell image with its corresponding displacement field and annotated cell type (C2C12) in panel (i). The reference traction force field obtained using Bayesian Fourier-transform traction cytometry (BFTTC) is shown in panel (ii). 
(C, i–iv)~Zoomed-in views of predicted traction force fields obtained from ViT, ViT+cell-type, \textit{ViT+UNet}, and \textit{ViT+UNet}+cell-type models within the blue box region indicated in panel (B-ii). 
(D)~Statistical comparison of normalized root mean square error (NRMSE; i) and correlation coefficients (ii) among the four methods: ViT, ViT+cell-type, \textit{ViT+UNet}, and \textit{ViT+UNet}+cell-type. Error bars represent standard deviations.
(E, i–iv)~Joint distributions of force magnitudes $p(F_\text{Ref}, F_\text{NN})$ binned for the deep learning models ViT, ViT with added cell-type input (ViT+cell-type), \textit{ViT+UNet}, and \textit{ViT+UNet} with added cell-type input (\textit{ViT+UNet}+cell-type).}
    \label{fig:cell_type}
\end{figure*}

\subsection{Adding Cell-Type Information as Input Data Enhances Prediction Specificity and Accuracy}
In this section, we investigate whether incorporating cell-type information as an additional input can improve the prediction of traction forces. To achieve this, we adopt the Mask Transformer approach~\cite{strudel2021segmenter}, which was originally developed for image segmentation tasks. Figure~(\ref{fig:cell_type}-A) illustrates the architecture of ViT+UNet with integrated cell-type information alongside the displacement field as input. In our dataset, four cell types are included: WT (Wild Type), C2C12, DMD (Duchenne Muscular Dystrophy), and Generic cells. The cell-type text information is first encoded as integer values ranging from 1 to 4 through index mapping. Each index is then converted into its corresponding cell-type embedding within a $D$-dimensional embedding space. This embedding vector is concatenated with the linear embedding matrix without adding positional encoding in the ViT module, as shown in Fig.~(\ref{fig:cell_type}-A). Subsequently, this cell-type embedding participates in the Transformer but not in the convolutional decoder, following a similar approach to that used in the Mask Transformer for image segmentation~\cite{strudel2021segmenter}. Note that the path of the displacement input within the network remains identical to that in the ViT+UNet architecture without cell-type information.

To quantify the inference of traction forces from displacement data with added cell-type information as input, we trained two new models that incorporate cell-type information—ViT+cell\_type and ViT+UNet+cell\_type—and compared them with the original ViT+UNet and ViT models without cell-type information described in the previous section. These models have comparable network sizes, and their training parameter settings are identical. Figure~\ref{fig:cell_type}(B) shows a representative cell image along with its corresponding displacement field and annotated cell type (C2C12) in panel (i), while the reference traction force field obtained using Bayesian Fourier-transform traction cytometry (BFTTC) is shown in panel (ii). 

Figure~\ref{fig:cell_type}(C, ii–v) illustrates the comparison of predicted zoomed-in traction force fields obtained from ViT, ViT+cell\_type, ViT+UNet, and ViT+UNet+cell\_type, respectively. We found that ViT+UNet in Fig.~\ref{fig:cell_type}(C, iv) and ViT+UNet+cell\_type in Fig.~\ref{fig:cell_type}(C, v) exhibit similar shape patterns and force directions to the reference shown in Fig.~\ref{fig:cell_type}(C, i). However, the magnitudes predicted by ViT+UNet+cell\_type match the reference more closely than those produced by ViT+UNet. Furthermore, we observed that the shape pattern, force direction, and force magnitude of both ViT (Fig.~\ref{fig:cell_type}(C, ii)) and ViT+cell\_type (Fig.~\ref{fig:cell_type}(C, iii)) are inferior to those of ViT+UNet. Nevertheless, the predictions from ViT+cell\_type demonstrate improved shape patterns and force magnitudes compared with those obtained from the baseline ViT model.

Next, we plotted the joint distributions of force magnitudes between the reference and predicted traction forces—$p(F_\text{Ref}, F_\text{NN})$—for all four methods: ViT, ViT+cell\_type, ViT+UNet, and ViT+UNet+cell\_type, as shown in Fig.~\ref{fig:cell_type}(E, i–iv), respectively. We found that the two distributions corresponding to ViT+UNet—with and without cell-type information—are narrower and lie closer to the diagonal dotted line compared with those from the ViT models (with and without cell-type information). Moreover, the distributions of ViT+UNet+cell\_type and ViT+cell\_type are even narrower and closer to the diagonal dotted line compared with their respective counterparts, ViT+UNet and ViT.

Finally, we investigated the statistical comparison of NRMSE and correlation coefficients among the four methods—ViT, ViT+cell\_type, ViT+UNet, and ViT+UNet+cell\_type—using results from 34 test cases, as shown in Fig.~\ref{fig:cell_type}(D, i–ii). We found that incorporating cell-type information in both ViT+UNet and ViT significantly reduces NRMSE values to $0.4644\pm0.2248$ and $0.5676\pm0.1308$, respectively, compared with $0.4926\pm0.3142$ and $0.6722\pm0.2564$ obtained without cell-type information (Fig.~\ref{fig:cell_type}(D, i)). However, the correlation is not strongly affected by the inclusion of cell-type information for either ViT or ViT+UNet (Fig.~\ref{fig:cell_type}(D, ii)).

\section{Conclusion}
In this study, we developed \textit{ViT+UNet}, a hybrid deep learning architecture that integrates the local feature extraction capabilities of the classical U-Net with the global contextual awareness of Vision Transformers (ViT). By combining convolutional operations with transformer-based attention mechanisms, the proposed model effectively captures both fine-grained spatial features and long-range dependencies within traction force microscopy (TFM) data. As a result, \textit{ViT+UNet} achieves superior performance in reconstructing complex traction force fields compared with standalone U-Net and ViT architectures. This synergy between convolutional and transformer modules enables more accurate mapping from displacement fields to cellular traction forces, providing improved generalization across diverse experimental conditions.

Through systematic evaluation, we demonstrated that the \textit{ViT+UNet} hybrid model achieves high accuracy and robustness across diverse spatial scales while maintaining reliable inference under varying noise levels. Furthermore, incorporating cell-type information as auxiliary input enhances prediction specificity and accuracy without increasing network complexity. These findings emphasize the effectiveness of integrating convolutional and transformer-based mechanisms to simultaneously capture fine-grained local structures and long-range contextual dependencies within complex biophysical datasets. By bridging these complementary feature extraction paradigms, \textit{ViT+UNet} provides a unified framework capable of learning hierarchical representations that are both spatially precise and contextually aware—an essential requirement for accurate traction force reconstruction in heterogeneous biological environments.

Beyond its predictive performance, \textit{ViT+UNet} offers a flexible framework for integrating metadata into traction force microscopy analysis. The capability to include experimental variables such as cell type or imaging conditions enables more context-aware predictions that can generalize across heterogeneous datasets. Moreover, this approach provides a foundation for developing multimodal learning pipelines capable of fusing image data with biochemical or mechanical metadata to achieve holistic mechanobiological insights.

Although our current implementation focuses on two-dimensional TFM data, the underlying principles can be extended to three-dimensional systems to reconstruct volumetric traction forces in collagen matrices or tissue scaffolds. Future work may also incorporate temporal modeling using recurrent or attention-based modules to analyze dynamic cellular processes such as migration~\cite{hockenberry2025measurement} or wound healing~\cite{brugues2014forces} over time. Furthermore, coupling uncertainty quantification techniques with our hybrid architecture could yield interpretable confidence maps that assist researchers in assessing prediction reliability at different spatial regions~\cite{kandasamy2025uncertainty}.

Overall, our findings establish \textit{ViT+UNet} as an effective and extensible framework for traction force microscopy analysis. The proposed approach not only enables high-fidelity reconstruction of cellular traction forces but also provides adaptability to diverse experimental conditions through metadata-driven learning. We anticipate that these advancements will contribute toward more precise mechanobiological studies and inspire new research directions at the intersection of computational biomechanics, machine learning, and cell biology.


\bibliographystyle{IEEEtranN}  
\bibliography{reference}


 






\vfill

\end{document}